\documentclass[10pt,twocolumn,letterpaper]{article}

 \usepackage[pagenumbers]{cvpr} %

\definecolor{cvprblue}{rgb}{0.21,0.49,0.74}
\usepackage[pagebackref,breaklinks,colorlinks,allcolors=cvprblue]{hyperref}

\usepackage{booktabs}
\usepackage{multirow}
\usepackage{multicol}
\usepackage{color, colortbl}
\usepackage{pifont}%
\usepackage{array}

\def\paperID{6697} %
\def\confName{CVPR}
\def\confYear{2025}

\title{GPD-1: Generative Pre-training for Driving}

\author{Zixun Xie$^{1,}$\footnotemark[1] \quad 
Sicheng Zuo$^{2,}$\footnotemark[1] \quad 
Wenzhao Zheng$^{2,}$\footnotemark[1] $^{,}$\footnotemark[2] \quad 
Yunpeng Zhang$^3$ \quad 
Dalong Du$^{2,3}$ \quad \\
{Jie Zhou}$^2$ \quad 
{Jiwen Lu}$^2$ \quad
{Shanghang Zhang}$^{1,}$\footnotemark[3] \\
$^1$Peking University \quad 
$^2$Tsinghua University \quad 
$^3$Phigent Robotics \\
\texttt{xinfei\_21@hotmail.com; wenzhao.zheng@outlook.com} \\
Project Page: \url{https://wzzheng.net/GPD} \\
Large Driving Models: \url{https:/github.com/wzzheng/LDM}
}

\begin{document}

\twocolumn[{%
\renewcommand\twocolumn[1][]{#1}%
\vspace{-2mm}
\maketitle
\vspace{-9mm}
\begin{center}
    \centering
    \includegraphics[width=\linewidth]{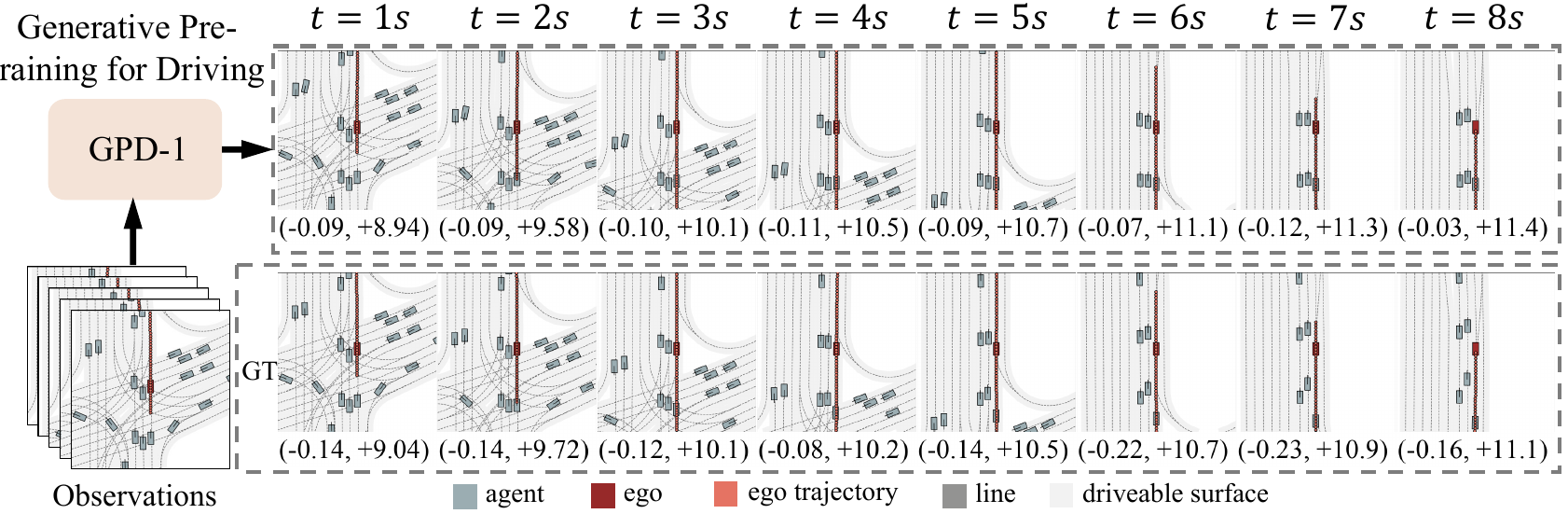}
    \vspace{-7mm}
    \captionof{figure}{
Given past 2D BEV observations, our pre-trained GPD-1 model can jointly predict future scene evolution and agent movements. 
This task requires both spatial understanding of the 2D scene and temporal modeling of how driving scenarios progress. 
We observe that GPD-1 successfully forecasts the movements of surrounding agents and future map elements.
Remarkably, it even generates more plausible drivable areas than the ground truth, showcasing its capacity to understand the scene rather than merely memorizing training data. 
However, it struggles to anticipate new vehicles entering the field of view, which is challenging due to their absence in the input data.
    }
\label{teaser}
\end{center}%
}]

\renewcommand{\thefootnote}{\fnsymbol{footnote}}
\footnotetext[1]{Equal contributions. $\dagger$Project leader. $\ddagger$Corresponding author.}
\renewcommand{\thefootnote}{\arabic{footnote}}

\begin{abstract}
Modeling the evolutions of driving scenarios is important for the evaluation and decision-making of autonomous driving systems.
Most existing methods focus on one aspect of scene evolution such as map generation, motion prediction, and trajectory planning.
In this paper, we propose a unified Generative Pre-training for Driving (GPD) model to accomplish all these tasks altogether without additional finetuning.
We represent each scene with ego, agent, and map tokens and formulate autonomous driving as a unified token generation problem.
We adopt the autoregressive transformer architecture and use a scene-level attention mask to enable intra-scene bi-directional interactions.
For the ego and agent tokens, we propose a hierarchical positional tokenizer to effectively encode both 2D positions and headings.
For the map tokens, we train a map vector-quantized autoencoder to efficiently compress ego-centric semantic maps into discrete tokens.
We pre-train our GPD on the large-scale nuPlan dataset and conduct extensive experiments to evaluate its effectiveness.
With different prompts, our GPD successfully generalizes to various tasks without finetuning, including scene generation, traffic simulation, closed-loop simulation, map prediction, and motion planning.
Code: \url{https://github.com/wzzheng/GPD}.
\end{abstract}
    
\section{Introduction}
\label{sec:intro}
Autonomous driving simulators~\cite{carla, Deepdrive, highway-env, kothari2021drivergym, amini2022vista, nvidia2020drivesim, caesar2022nuplanclosedloopmlbasedplanning} play a crucial role in developing and validating driving systems, enabling safe testing across various driving scenarios, including perception~\cite{lss, huang2021bevdet, li2022bevformer}, motion prediction~\cite{gu2021densetnt, zhou2023query, seff2023motionlm}, and trajectory planning~\cite{zheng2023occworldlearning3doccupancy,cheng2023rethinkingimitationbasedplannerautonomous, uniad, jiang2023vad, chen2024vadv2, li2024hydra-mdp, zheng2024genad}.

Typical components of the driving simulators can include scene generation, traffic simulation, closed-loop simulation, and motion planning. Particularly, recent advancements in BEV (bird's eye view) representations have demonstrated the feasibility of using simulators to replicate real-world driving conditions and challenges \cite{chitta2024sledgesynthesizingdrivingenvironments}.
Such simulators have become essential for testing complex behaviors, understanding interaction dynamics, and ensuring robustness against potential failures, thus contributing to safe and reliable autonomous driving systems.
However, existing methods for scene evolution in autonomous driving are generally specialized and limited to specific aspects of the simulator, such as map generation~\cite{PGDrive, Adversarial_PCG, hartmann2017streetgan}, motion prediction~\cite{suo2021trafficsim, social_pooling, yuan2021agentformer, chitta2024sledgesynthesizingdrivingenvironments}, or trajectory planning \cite{chitta2024sledgesynthesizingdrivingenvironments}.
Considering these approaches typically focus on one isolated task, there exists no unified framework that integrates these aspects into a cohesive model for holistic simulation.
For example, the recent method SLEDGE~\cite{chitta2024sledgesynthesizingdrivingenvironments} is only trained to reconstruct single frames and lacks control interfaces, limiting its ability to support various downstream tasks.
They cannot fully leverage the scene-level information including the temporal evolution across scene elements and the interactions between dynamic agents and map elements, making it challenging to generalize to different downstream tasks.

In this paper, we propose to unify these elements with a \textbf{G}enerative \textbf{P}re-training for \textbf{D}riving (GPD-1) model. We encode the map, agents, and ego vehicle as a unified set of tokens, enabling us to formulate the scene evolution as the generative prediction of the scene tokens.
We adopt an autoregressive transformer architecture with a scene-level attention mask that enables bi-directional interactions within the scene, allowing the model to efficiently capture dependencies among the ego, agent, and map tokens. 
For ego and agent tokens, we propose a hierarchical positional tokenizer, which effectively encodes the BEV positions and headings. The positional tokenizer transforms the continuous agent positions into discrete tokens, which significantly reduces the noise in feature space. 
For map tokens, we leverage a vector-quantized autoencoder (VQ-VAE)~\cite{vqvae} to compress ego-centric semantic maps into discrete tokens. By representing map information through discrete tokens, we eliminate the complexity of predicting continuous map coordinates, simplifying the learning process and enhancing generalization.
To demonstrate the effectiveness of our GPD-1 model, we conduct a series of challenging experiments across diverse tasks. Our model, as shown in Figure \ref{teaser}, without any fine-tuning, is capable of performing scene generation, traffic simulation, closed-loop simulation, and motion planning. Specifically, scene generation involves initializing a scene and allowing the model to generate agent, map, and ego information smoothly. Traffic simulation provides the ground-truth map and initial agent states, with the model predicting the evolution of subsequent frames. Closed-loop simulation, given a ground-truth map and ego trajectory, allows the model to dynamically adapt agent trajectories in response to ego movements. Finally, for motion planning, the model generates ego trajectories in response to the provided agent and map information. 
With further fine-tuning, GPD-1 can achieve state-of-the-art performance on downstream tasks, particularly the motion planning task from the nuPlan benchmark.
\section{Related Work}
\label{sec:formatting}

\textbf{Discrete Tokens for Autonomous Driving.} 
Tokenized discrete representations have become popular for capturing complex spatial layouts with efficiency and interpretability. 
VQ-VAE \cite{vqvae} introduced a codebook mechanism to construct an encoder-decoder architecture within a tokenized discrete latent space, enabling richer, more compact representations of high-dimensional data. VQ-VAE-2~\cite{vq-vae2} further enhanced the framework with hierarchical quantized codes and autoregressive priors.
Following this direction, models like VQ-GAN \cite{esser2021tamingtransformershighresolutionimage}, DALL-E \cite{ramesh2021zeroshottexttoimagegeneration}, and VQ-Diffusion \cite{gu2022vectorquantizeddiffusionmodel} map inputs into discrete tokens corresponding to codebook entries, allowing simplified yet expressive representations. 
Recent works in visual pre-training \cite{bao2022beitbertpretrainingimage, peng2022beitv2maskedimage} employ similar tokenization strategies, using tokens to represent image patches and predicting masked tokens as a proxy task to enhance model robustness and versatility. 
To represent the map elements, recent methods on map reconstruction~\cite{liu2023vectormapnet, liao2022maptr} and end-to-end driving~\cite{jiang2023vad} encoding each map element into a vectorized representation for modeling, which ignores the scene-level structures.

We apply tokenizing to BEV-based autonomous driving scenarios and encode map features into discrete tokens. 
Our method addresses common issues in BEV modeling, such as computational inefficiencies and inconsistency in representations, by minimizing spatial noise and providing a unified structure for map and agent information.

\textbf{Data-Driven Autonomous Driving Simulation.} 
Traditional simulation techniques often involve replaying logged driving data to emulate various driving conditions \cite{caesar2022nuplanclosedloopmlbasedplanning, fong2021panopticnuScene, gulino2023waymaxaccelerateddatadrivensimulator, karnchanachari2024learningbasedplanningthenuplanbenchmark}. 
For instance, conventional simulators like nuPlan \cite{caesar2022nuplanclosedloopmlbasedplanning} rely heavily extensive driving logs to cover diverse scenarios. However, these simulations demand massive storage capacities, making them resource-intensive and challenging for broader accessibility. Also, these model-driven simulators require complicated rule-based modules for scene generation, agent behaviors, and rendering.
To this end, data-driven simulation methods are proposed for sensor rendering~\cite{zhou2024drivinggaussian, yan2024streetgs, yu2024sgd, s3gaussian}, road network generation~\cite{PGDrive, Adversarial_PCG, hartmann2017streetgan}, and agent behavior prediction~\cite{suo2021trafficsim, social_pooling, yuan2021agentformer, chitta2024sledgesynthesizingdrivingenvironments}.
For example, SLEDGE \cite{chitta2024sledgesynthesizingdrivingenvironments} leverages generative models to simulate scenes with compact vectorized data, enabling efficient use of storage without compromising on scenario diversity or complexity. 
While effective, they lack adaptability in dynamically modeling interactions between agents and the surrounding map, limiting their application for reactive tasks. 
Differently, our framework aims to bridge this gap by incorporating a generative model capable of scene evolution and thus allows for interactive and flexible scene generation that supports various downstream tasks.

\section{Proposed Approach}

\subsection{2D Map Scene Tokenizer}
A key aspect of autonomous driving is capturing spatial information about the environment accurately and efficiently. 
To achieve this, we employ a 2D Map Scene Tokenizer that transforms complex, vector-based map representations into discrete tokens, which can be effectively modeled within a generative framework. 
This tokenizer is designed to simplify the continuous spatial features into a structured, discrete format, enabling our model to incorporate map information seamlessly alongside agent and ego tokens.

\textbf{Map Vector Rasterization.} The map data consists of vector representations of lines, each defined by multiple points.
Directly encoding these vectors poses challenges due to the lack of spatial relationships within the vector formats.
To resolve this, we rasterize the map vectors into a 2D canvas centered at the ego vehicle and only represent the immediately visible region.
This rasterized map is represented as a binary image $I \in \mathbb{R}^{H \times W}$, where the interpolated line segments and background regions are marked as 1 and 0.

\textbf{Feature Extraction and Quantization.} 
To efficiently represent the map data, we use a vector-quantized autoencoder (VQ-VAE)~\cite{vqvae} that converts continuous map features into discrete tokens. 
The rasterized map $I$ is first encoded by ResNet-50 \cite{he2015deepresiduallearningimage} into compact features $\hat{z} \in \mathbb{R}^{H/d \times W/d \times C}$, where $H = W = 256$, $d$ is the downsampling factor, and $C$ is the feature dimension. 
For quantization, we introduce a codebook $V \in \mathbb{R}^{K \times D}$ with $K$ discrete codes, each capturing a high-level feature of the scene. Each map feature $\hat{z}_{ij}$ in $\hat{z}$ is quantized by mapping it to the nearest code in $V$:
\begin{equation}
z_q = Q(z_c) = \underset{v_k \in V}{\arg \min} \| \hat{z}_{ij} - v_k \|_2,
\end{equation}
where $\| \cdot \|_2$ denotes the L2 norm. 
Here, \( Q(z_c) \) represents the quantization function that maps the continuous latent vector \( z_c \) to its nearest neighbor in the codebook \( V \), resulting in the discrete representation \( z_q \). 
These tokens provide a compact and consistent representation of the map information and encode spatial structure while reducing model complexity.

\textbf{Reconstruction with Discrete Queries.} 
We follow the DETR \cite{carion2020end} decoding approach defined in SLEDGE \cite{chitta2024sledgesynthesizingdrivingenvironments} to decode the quantized map tokens into the Vector Lane Representation as outlined in SLEDGE. 
For aligning the generated and ground-truth map lines, we also adopt the Hungarian algorithm for matching, using the same supervision loss setup as SLEDGE to ensure accurate map reconstruction.
The map tokenizer transforms vector-based maps into compact discrete space, encoding essential spatial relationships. This representation facilitates the modeling of dynamic scene elements within the generative framework.

\begin{figure}[t]
\centering
\includegraphics[width=0.475\textwidth]{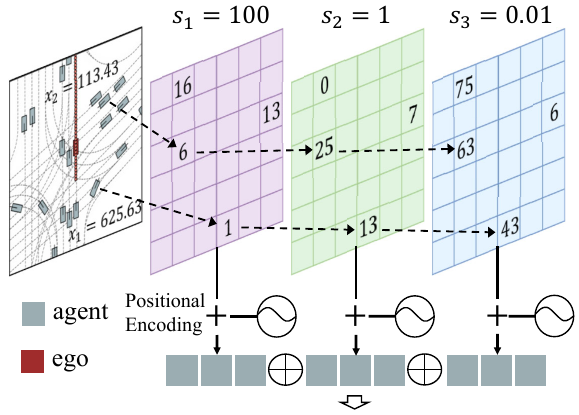}
\vspace{-7mm}
\caption{\textbf{Illustration of the agent tokenizer.} 
We use a set of thresholds to categorize agent states into different ranges to convert continuous information into discrete representations.
}
\label{fig:tokenizer}
\vspace{-6mm}
\end{figure}

\subsection{Agent Tokenizer}
In autonomous driving simulations, accurately representing dynamic agents within the scene is essential for realistic and coherent scene generation. To efficiently encode agent data, we introduce a hierarchical positional tokenizer to capture both spatial (2D position) and angular (heading) information. This tokenizer enables the model to represent complex agent dynamics while reducing the feature space, making the generative process more manageable.

\textbf{Multi-Level Quantization.} Each agent coordinate, denoted as a general variable \( p \) (e.g., \( x \), \( y \), or heading), undergoes multi-level quantization across $N$ hierarchical levels, represented by a set of thresholds $\{s_1, s_2, \dots, s_N\}$, where each $s_i$ denotes a specific scale of granularity.

For the first level, the quantized value \( q_1 \) is calculated as:
\begin{equation}
q_1 = \text{floor}\left(\frac{p}{s_1}\right).
\end{equation}

For levels \( i > 1 \), the quantization is performed on the residual after accounting for the previous levels:
\begin{equation}
q_i = \text{floor}\left(\frac{p - \sum_{j=1}^{i-1} q_j \cdot s_j}{s_i}\right).
\end{equation}

\begin{figure*}[t]
\centering
\includegraphics[width=0.9\textwidth]{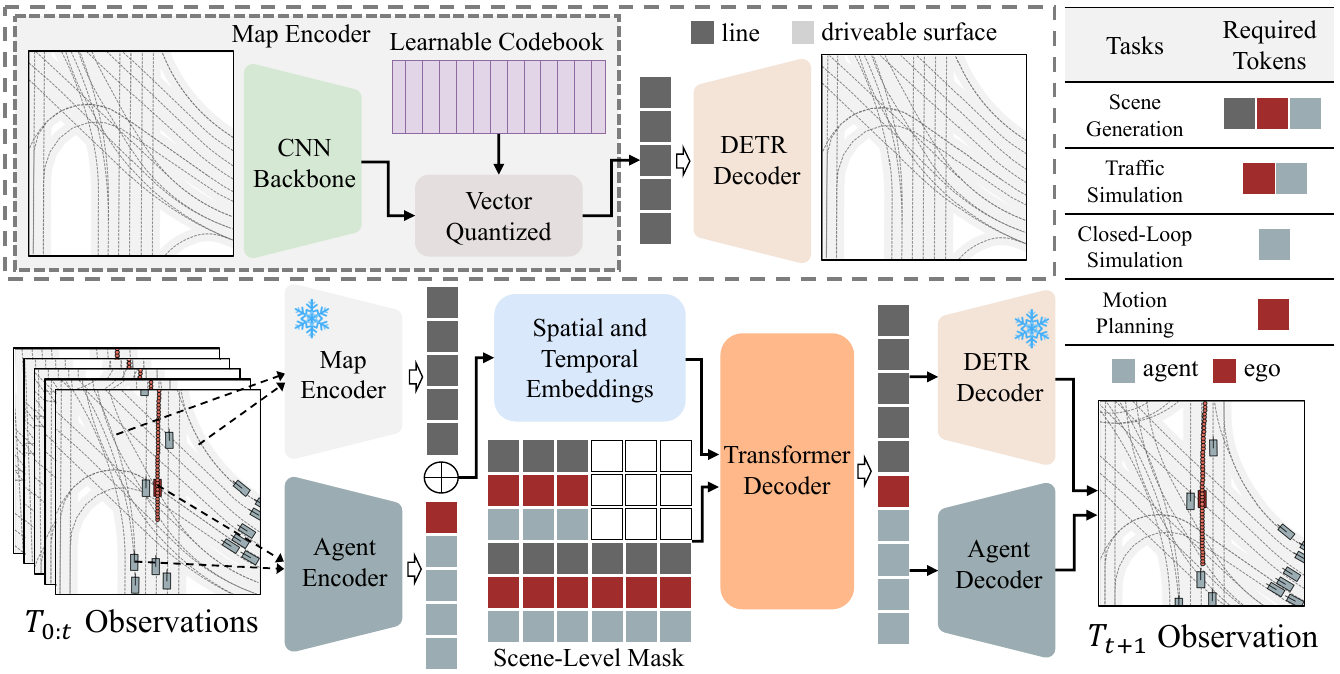}
\vspace{-3mm}
\caption{\textbf{Framework of our GPD-1 model for 2D scene forecasting and motion planning.} Our model adapts the GPT-like architecture for autonomous driving scenarios with two key innovations: 1) a 2D map scene tokenizer that generates discrete high-level representations of the 2D BEV map, and 2) a hierarchical quantization agent tokenizer to encode agent information. Using a scene-level mask, the autoregressive transformer predicts future scenes by conditioning on both ground truth and previously predicted scene tokens during training and inference, respectively.
}
\label{fig:framework}
\vspace{-5mm}
\end{figure*}

This iterative quantization ensures that each level captures progressively finer details by focusing on the residual not captured by previous levels. The result is a set of $N$ quantized values \( \{q_1, q_2, \dots, q_N\} \), each representing the coordinate at different levels of precision.

\textbf{Positional Embedding.} After quantization, we incorporate a fixed sinusoidal embedding to each quantized level, capturing its relative position within the feature space. This sinusoidal encoding is based on the classic positional encoding introduced in Transformers \cite{kazemian2024attentionneedoptimizewind}, which provides spatial context and preserves positional relationships within the discrete embedding space. The embedding for each quantized level is defined as:
\begin{equation}
e_i = \text{SinusoidalPositionEncoding}(q_i),
\end{equation}
where $e_i$ is the embedding corresponding to the quantized value \( q_i \). Finally, positional embeddings $\{e_1, e_2, \dots, e_N\}$ from all quantized levels are concatenated to form the final positional encoding vector for each coordinate:
\begin{equation}
\text{pos\_vec} = e_1 \oplus e_2 \oplus \dots \oplus e_N,
\end{equation}
where $\oplus$ denotes concatenation. This results in a comprehensive, multi-level representation for the agent coordinate \( p \), capturing both fine and coarse spatial details.

This hierarchical tokenization process is applied uniformly to \( x \), \( y \), and heading values, providing a consistent approach to encoding spatial and angular information for each agent. The combined embeddings are then concatenated and passed through an MLP \cite{haykin1994neural} to map them to the specified model dimension. For agents that are outside the visible area, we apply a unified set of learnable parameters, allowing the model to autonomously learn representations for unseen agents.

The agent tokenizer in Figure \ref{fig:tokenizer} transforms agent positions and headings into discrete embeddings, enabling a structured representation of spatial and angular relationships. This tokenization reduces positional noise and introduces consistency in feature space, improving the ability to learn and predict agent dynamics effectively.

\subsection{Generative Transformer for Scene Modeling}

In autonomous driving, the ability to model the evolution of an entire scene is essential for predicting dynamic interactions among agents and understanding future outcomes. We employ an autoregressive transformer architecture to handle scene modeling, inspired by the sequential generation framework of GPT \cite{brown2020language}. Our approach incorporates a scene-level attention mask that enables bi-directional interactions among tokens within each frame, allowing for a comprehensive understanding of both spatial and temporal relationships, illustrated in Figure \ref{fig:framework}.

Each scene, corresponding to a single frame, consists of a fixed number of map tokens and agent tokens. The map tokens originate from the 2D Map Scene Tokenizer as discrete latent representations $z_q$ obtained via VQ-VAE, and their quantity is determined by the dimensionality of the latent space. The agent tokens, produced by the agent tokenizer, represent individual agents within the scene, with a fixed number assigned to each frame.

\textbf{Spatial and Temporal Embeddings.} To provide the model with structured information about the spatial layout and temporal progression, we add learnable spatial and temporal embeddings. The spatial embedding associates each token with its role as either a map or agent token, ensuring that the model understands the distinct functions of each element within the scene. The temporal embedding encodes the sequence order across frames, capturing the progression of events over time. These embeddings allow the model to maintain a consistent structure, where each frame is composed of a fixed arrangement of map and agent tokens, facilitating the understanding of spatial relationships and temporal dependencies across frames.

\textbf{Scene-Level Attention Mask.} The attention mechanism uses a scene-level attention mask, \( M \), to control interactions within and across tokens in a frame. The mask \( M \) has dimensions $[T_{\text{max}} \cdot N, T_{\text{max}} \cdot N]$, where \( T_{\text{max}} \) is the maximum number of time steps, and \( N = N_{\text{agent}} + N_{\text{map}} \) represents the total number of agent and map tokens in each frame.

Initially, the mask is set as an upper triangular matrix to prevent tokens from attending to future frames, enforcing an autoregressive structure. Additionally, for each time step \( t \), the mask is adjusted to allow full interaction among tokens within the same frame, defined by:
\begin{equation}
M[t \cdot N : (t + 1) \cdot N, \; t \cdot N : (t + 1) \cdot N] = 0.
\end{equation}
This configuration allows for intra-frame spatial interactions among map and agent tokens within a single time step while blocking information flow from future frames.

\textbf{Autoregressive Modeling.} Following the architecture of GPT, our transformer decoder processes each scene in an autoregressive manner, predicting the evolution of scene tokens over time. At each time step, the decoder receives the spatially and temporally embedded scene tokens, processes them with the scene-level attention mask, and predicts the next set of tokens. This can be formulated as:
\begin{equation}
\hat{T}_{t+1} = \text{TransformerDecoder}(T_{0:t}, M),
\end{equation}
where \( T_{0:t} \) denotes the set of tokens during time steps \( 0 \) to \( t \), and \( M \) is the scene-level attention mask. 
This learns both the spatial relationships among tokens within a frame and the temporal dependencies across frames, which is crucial for generating realistic and dynamic driving scenes.

The generative transformer leverages a structured combination of map and agent tokens, enriched by spatial and temporal embeddings, to predict scene evolution. The scene-level attention mask enables nuanced interactions within each frame, enhancing the ability to learn coherent spatial relationships and temporal progression, making it highly suitable for autonomous driving scenarios.

\begin{table*}[t!]
    \centering
    \caption{\textbf{Applications of the proposed GPD-1 on various tasks.}
}
\vspace{-3mm}
\setlength{\tabcolsep}{5pt}
    \begin{tabular}{c|ccc|ccc|cccc}
\toprule
        \multirow{2}{*}{Predicted Duration} & \multicolumn{3}{c|}{Ego Trajectory}  & \multicolumn{3}{c|}{Agent Trajectory}  & \multicolumn{3}{c}{Map} \\
        & ADE$\downarrow$ & FDE $\downarrow$ & Coll. (\%) $\downarrow$ & ADE $\downarrow$ & FDE $\downarrow$ & Coll. (\%) $\downarrow$ & F1 $\uparrow$ & Lat. $\downarrow$ & Ch. $\downarrow$ \\ 
        \midrule
        \textbf{Scene Generation}: \\
        3$s$ & 0.662 & 1.625 & 0.000 & 0.613 & 1.962 & 0.533 & 0.798 & 0.196 & 3.540 \\ 
        5$s$ & 1.539 & 4.127 & 0.300 & 1.172 & 4.588 & 0.764 & 0.665 & 0.208 & 8.792 \\ 
        8$s$ & 3.509 & 9.816 & 2.910 & 2.126 & 7.765 & 1.465 & 0.552 & 0.210 & 17.83 \\ \midrule
        \textbf{Traffic Simulation}: \\
        3$s$ & 0.631 & 1.718 & 1.817 & 0.572 & 1.530 & 0.423 & - & - & - \\ 
        5$s$ & 1.643 & 4.714 & 3.034 & 1.110 & 3.661 & 0.742 & - & - & - \\ 
        8$s$ & 4.001 & 11.27 & 3.792 & 2.201 & 7.438 & 1.243 & - & - & - \\ 
      \midrule
        \textbf{Close-Loop Simulation}: \\
        3$s$ & - & - & - & 0.610 & 1.820 & 0.530 & - & - & - \\ 
        5$s$ & - & - & - & 1.133 & 4.284 & 0.806 & - & - & - \\ 
        8$s$ & - & - & - & 1.916 & 6.817 & 1.271 & - & - & - \\ 
     \midrule
        \textbf{Motion Planning}: \\
        3$s$ & 0.645 & 1.720 & 0.600 & - & - & - & - & - & - \\ 
        5$s$ & 1.627 & 4.560 & 1.446 & - & - & - & - & - & - \\ 
        8$s$ & 3.813 & 10.47 & 3.749 & - & - & - & - & - & - \\ 
        \bottomrule
    \end{tabular}
    \vspace{-5mm}
\label{tab:main}
\end{table*}

\begin{table*}
\vspace{6pt}
\centering
\setlength{\tabcolsep}{10pt}
\renewcommand{\arraystretch}{1}
\caption{\textbf{Motion planning performance on nuPlan.} 
}\vspace{-3mm}
\begin{tabular}{cc|ccc|ccc}
\toprule
\multicolumn{2}{c}{Model} & \multicolumn{3}{c}{Test14-random} & \multicolumn{3}{c}{Test14-hard} \\ \midrule
Method & \multicolumn{1}{l|}{Configuration} & OLS & NR-CLS & \multicolumn{1}{c|}{R-CLS} & OLS & NR-CLS & R-CLS \\ \midrule
\multirow{2}{*}{planTF \cite{cheng2023rethinkingimitationbasedplannerautonomous}} & \multicolumn{1}{l|}{w/ history + shared encoder} & 90.20 & 56.50 & \multicolumn{1}{c|}{56.28} & \textbf{88.25} & 48.60 & \textbf{51.32} \\
 & \multicolumn{1}{l|}{w/ history + separate encoder} & \textbf{90.28} & 61.02 & \multicolumn{1}{c|}{59.85} & 86.77 & \textbf{51.98} & 49.34 \\ \midrule
\multirow{2}{*}{GPD-1} & \multicolumn{1}{l|}{w/ history + wo/ pretrain} & 29.63 & 13.46 & \multicolumn{1}{c|}{12.93} & 21.52 & 10.05 & 9.31 \\
 & \multicolumn{1}{l|}{w/ history + w/ pretrain} & 87.05 & \textbf{63.45} & \multicolumn{1}{c|}{\textbf{63.52}} & 81.68 & 47.92 & 46.69 \\ \bottomrule
\end{tabular}
\label{plan1}
\vspace{-7mm}
\end{table*}

\subsection{GPD-1: Generative Pre-training for Driving}

Our Generative Pre-training for Driving (GPD-1) model uses a two-stage training process to build a robust foundation for autonomous driving simulations and planning tasks.
We first train the Map VQ-VAE latent tokenizer, adopting the L1 error for map line position and binary cross-entropy (BCE) to assess map line visibility, as defined in SLEDGE \cite{chitta2024sledgesynthesizingdrivingenvironments}. Additionally, to improve codebook stability and precision, we include the mean squared error (MSE) loss to encourage accurate quantization. This stage creates a high-fidelity map latent space that accurately encodes spatial structure, forming a solid base for scene generation.

In the second stage, the trained map tokenizer is frozen and used to extract latent representations of the map for each frame, which serve as both inputs and ground truth for further training. Cross-entropy (CE) loss is used to match generated tokens with their correct codebook entries, ensuring accurate map reconstruction. We treat both ego and agent tokens equally, using smooth L1 loss to calculate positional errors and BCE loss for binary classification of presence. This structured training allows the model to capture both spatial and temporal scene dynamics, enabling consistent scene modeling across diverse scenarios.

GPD-1 allows it to perform a wide array of downstream tasks without additional fine-tuning, demonstrating flexibility across critical autonomous driving applications.

\textbf{Scene Generation}: GPD-1 autonomously generates complete scenes by initializing a scene setup and predicting the spatial and temporal evolution of agents, the ego vehicle, and map features. This task is essential for creating diverse driving scenarios from minimal initial inputs.
    
\textbf{Traffic Simulation}: By initializing the model with a ground-truth map and initial agent states, GPD-1 accurately predicts how traffic evolves across frames. This simulation capability is crucial for evaluating and training autonomous driving models in dynamic environments, where understanding the flow of traffic is fundamental.
    
\textbf{Closed-Loop Simulation}: Given a ground-truth map and ego trajectory, the model can dynamically adapt agent behaviors in response to the ego vehicle's movements. This setup aligns closely with the closed-loop interactive settings in the nuPlan Challenge \cite{caesar2022nuplanclosedloopmlbasedplanning}, where agent reactions to the ego behavior are generated through the model rather than relying on conventional rule-based algorithms.
    
\textbf{Motion Planning}: GPD-1 supports ego trajectory planning by generating routes in response to a given set of agent and map information. This planning capacity closely aligns with practical autonomous driving needs, offering a data-driven alternative to conventional planning methods.
    
\textbf{Conditional Generation}: GPD-1 can also handle conditional generation, allowing users to define specific conditions such as initial agent trajectories, the number of agents, or vector-based map features. With these constraints, GPD-1 autonomously generates compatible scene evolutions, enabling simulation of targeted, scenario-specific driving conditions for fine-grained control.

\textbf{Enhanced Performance with Fine-Tuning.} Fine-tuning on specialized datasets or specific task scenarios further enhances the performance of GPD-1, especially in complex planning tasks. Fine-tuning enables GPD-1 to generate extended, precise trajectories that meet the rigorous standards of challenges such as the nuPlan Planning Challenge, where both closed-loop and open-loop performance are critical for accurate trajectory prediction.

The generative pre-training equips GPD-1 with a flexible, robust structure that accommodates a broad spectrum of tasks in autonomous driving. From scene generation to nuanced conditional simulations, GPD-1 serves as an adaptable and comprehensive solution for realistic, responsive driving simulations and trajectory planning, fulfilling essential needs in autonomous driving research and development.

\section{Experiments}

\subsection{Datasets}
We conducted extensive experiments on the nuPlan~\cite{caesar2022nuplanclosedloopmlbasedplanning} dataset.
nuPlan is a large-scale closed-loop planning benchmark designed for long-term decision-making evaluation for autonomous driving. 
It provides 1300 hours of driving data recorded from four different urban areas, which are divided into 75 distinct scenario types with automated labeling tools.
The data is collected with a vehicle with eight cameras providing a full $360^{\circ}$ horizontal field of view and a LiDAR sensor to obtain point cloud scans of the scenes.

\begin{figure*}[t]
\centering
\includegraphics[width=0.93\textwidth]{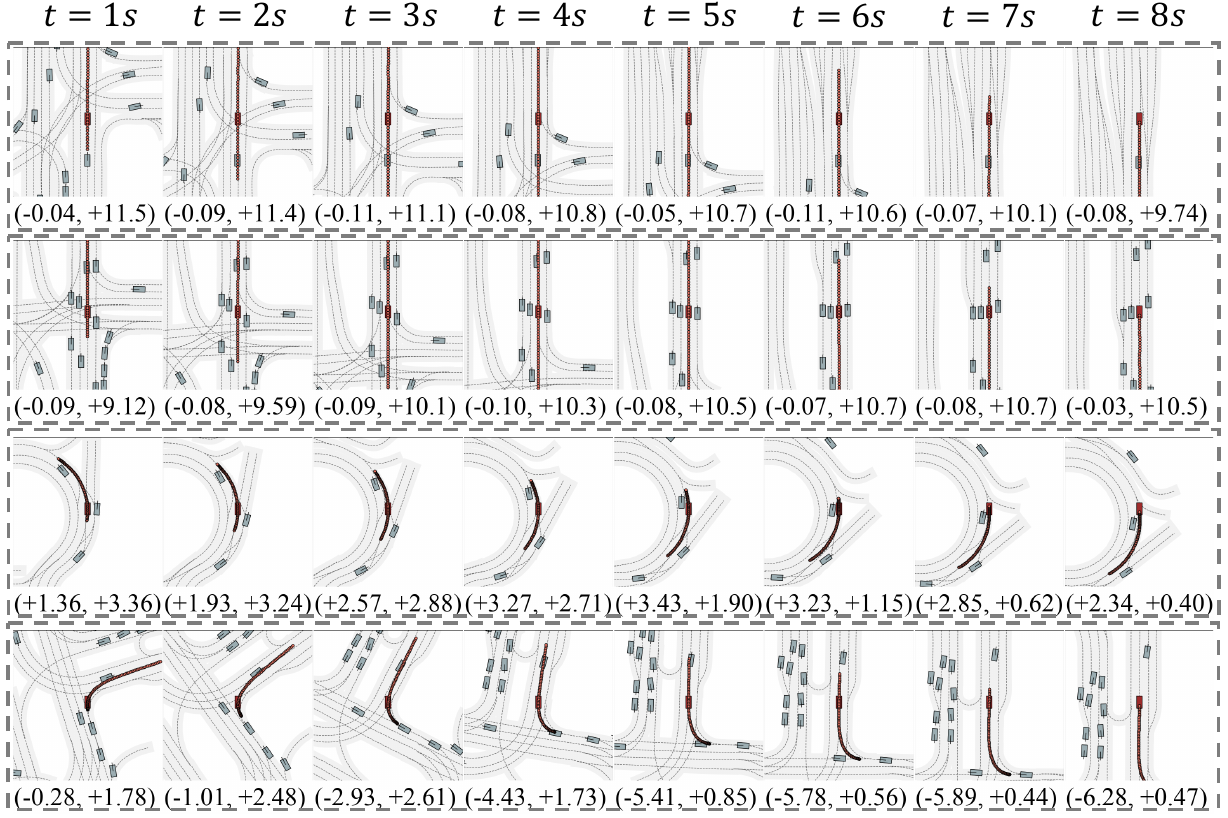}
\vspace{-4mm}
\caption{\textbf{Visualizations of the Scene Generation Task across different types of scenarios.}
}
\label{fig:vis}
\vspace{-7mm}
\end{figure*}

\subsection{Experimental Settings}
We employ the official evaluation metrics~\cite{caesar2022nuplanclosedloopmlbasedplanning} to evaluate the planning performance of our GPD-1, including the open-loop score (OLS), non-reactive closed-loop score (NR-CLS), and reactive closed-loop score (R-CLS). 
R-CLS and NR-CLS use the same calculation methods.
R-CLS includes background traffic control using an Intelligent Driver Model (IDM)~\cite{IDM} during simulations.
The closed-loop score is a composite score ranging from 0 to 100, which considers comprehensive factors such as traffic rule adherence, human driving resemblance, vehicle dynamics, goal attainment, and other metrics specific to the scenario. 
We include more implementation details in Section~\ref{app:imple}.

\subsection{Main Results}
\label{sec:MainResult}
To demonstrate the generality of GPD-1, we utilized it across multiple downstream tasks without any fine-tuning. As shown in Table~\ref{tab:main}, we present the model's performance across various settings. In these experiments, we provide a fixed 2-second map and agent data as initial information and use different prompt settings.

Overall, the autoregressive model performs best with fewer iterations. For instance, predicting 5 seconds into the future requires only 50 iterations and yields strong results. However, as the number of iterations increases, cumulative errors grow at an approximately quadratic rate.

\textbf{Scene Generation.}
The scene generation (SG) task setup is closest to the conditions in our training phase. The performance metrics for both the agents and the ego vehicle are similar, as the model treats the ego vehicle as an ordinary agent without any special adjustments. Figure~\ref{fig:vis} shows that even in complex scenarios (e.g., navigating sharp turns or congested areas), our ordinary agents maintain strong performance. This level of robustness is generally unattainable by traditional planning models like PlanTF.

\begin{table}[t]
\caption{\textbf{Performance on the map prediction task.} 
}\vspace{-3mm}
\centering
\setlength{\tabcolsep}{0.05\linewidth}
\begin{tabular}{l|ccc}
\toprule
Duration  & F1 $\uparrow$   & Lat. $\downarrow$  & Ch. $\downarrow$   \\
\toprule
w/ agents and ego\\
3$s$    & 0.874 & 0.203 & 2.250 \\
5$s$     & 0.828 & 0.223 & 3.332 \\
8$s$     & 0.770 & 0.234 & 6.516 \\
\toprule
w/ ego \\
3$s$     & 0.945 & 0.164 & 1.160 \\
5$s$    & 0.913 & 0.201 & 2.660 \\
8$s$     & 0.871 & 0.243 & 4.374 \\
\bottomrule
\end{tabular}
\label{map}
\vspace{-7mm}
\end{table}

\begin{figure*}[t!]
\centering
\includegraphics[width=0.93\textwidth]{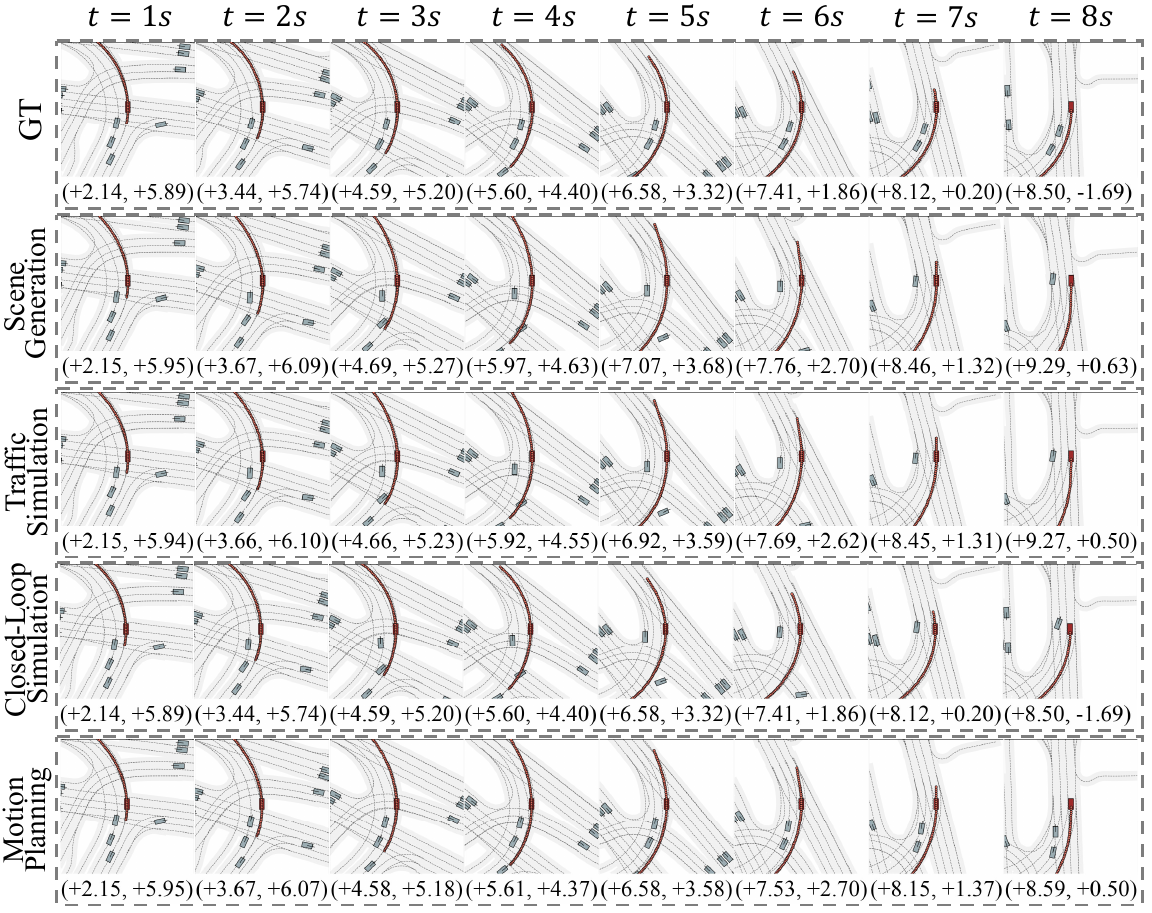}
\vspace{-4mm}
\caption{\textbf{Visualizations of the scene generation, traffic simulation, closed-loop simulation, and motion planning tasks.}
}
\label{vis2}
\vspace{-6mm}
\end{figure*}

\textbf{Traffic Simulation.}
We provided ground truth maps in this setting. 
The prediction error for the ego vehicle increases due to the cumulative error inherent in autoregressive models. Over extended time steps, the ego deviates increasingly from its original trajectory, while the map remains grounded in ground truth.

\textbf{Clopsed-Loop Simulation.}
For closed-loop simulation, the agents adapt to changes in the ego trajectory, maintaining a low collision rate and demonstrating strong reliability.

\textbf{Motion Planning.}
Motion planning is similar to the non-interactive closed-loop setting in nuPlan.
We used the model directly without fine-tuning or additional data augmentation, yet it still achieved commendable results.

\begin{table}[t]
\caption{\textbf{Effect of quantization.} 
We report performance on the generated trajectory quality of both the ego vehicle and agents.
}\vspace{-3mm}
\centering
\setlength{\tabcolsep}{0.009\linewidth}
\begin{tabular}{l|c|c}
\toprule
Method     & ADE (Ego) $\downarrow$ & ADE (Agents)  $\downarrow$ \\
\midrule
GPD-1 w Quantization   & \textbf{0.06}   & \textbf{0.26}   \\
GPD-1 w/o Quantization & 0.20   & 0.43  \\
\bottomrule
\end{tabular}
\label{ablation}
\vspace{-7mm}
\end{table}

\subsection{Results and Analysis}

\textbf{nuPlan Motion Planning Challenge.} 
The versatile representation enables seamless application to various downstream tasks, and even minimal fine-tuning can greatly enhance its performance on specific tasks. As shown in Table ~\ref{plan1}, we added only a single decoder layer to decode the ego token to meet the nuPlan challenge requirements. Without relying on complex data augmentation or post-processing techniques, our model achieves performance comparable to PlanTF and even surpasses it in certain metrics.

\textbf{Map Prediction.} In the map prediction experiment, we evaluated the model under two settings: 1) providing ground truth for both the agents and the ego vehicle to generate the map, and 2) providing only the ego ground truth and making all other agents invisible to generate the map. This experiment validates the conditional generation capability. As shown in Table~\ref{map}, the map prediction quality improves significantly when only the ego is given as input. This is because the map is centered on the current ego car, making it highly correlated with the state of the ego.

\textbf{Effect of Quantization.}
Table~\ref{ablation} demonstrates the impact of quantizing agent states on the per-frame performance of both the ego and agents. 
We see that the quantized discrete agent information combined with discretized maps jointly reduces the learning complexity of the feature space.

\textbf{Visualizations.}
Figure~\ref{fig:vis} shows the performance under the Scene Generation setting in complex scenarios. The results demonstrate that even in highly intricate road conditions, the map can be generated smoothly. In two turning scenarios, both the ego vehicle and agents follow a natural trajectory at a relatively steady speed. Similarly, in two straight-driving scenarios, the model effectively captures surrounding agents' actions (e.g., turning, driving, and decelerating) while maintaining a stable forward speed.

Figure~\ref{vis2} illustrates the performance in a more complex intersection-turning scenario across different settings. The quality of map generation is notably satisfactory, and for both agents and the ego vehicle, the performance closely matches the ground truth in all tasks, except where ground truth data is explicitly used. This consistency highlights the robustness of our model.

\section{Conclusion}
In this paper, we have introduced Generative Pre-training for Driving (GPD-1) for autonomous driving which models the joint evolution of ego movements, surrounding agents, and scene elements. We employ a hierarchical agent tokenizer and a vector-quantized map tokenizer to capture high-level spatial and temporal information, while an autoregressive transformer with scene-level attention predicts future scenarios across multiple driving tasks.
Extensive results demonstrate that GPD-1 effectively generalizes to diverse tasks, such as scene generation, traffic simulation, and motion planning, without additional fine-tuning. We believe that GPD-1 represents a foundational step toward a fully integrated, interpretable framework for autonomous driving.

\appendix

\begin{table*}[t!]
    \centering
    \caption{\textbf{Applications of the proposed GPD-1 on various tasks in the Test14-hard setting.}
}
\vspace{-3mm}
\setlength{\tabcolsep}{5pt}
    \begin{tabular}{c|ccc|ccc|cccc}
\toprule
        \multirow{2}{*}{Predicted Duration} & \multicolumn{3}{c|}{Ego Trajectory}  & \multicolumn{3}{c|}{Agent Trajectory}  & \multicolumn{3}{c}{Map} \\
        & ADE$\downarrow$ & FDE $\downarrow$ & Coll. (\%) $\downarrow$ & ADE $\downarrow$ & FDE $\downarrow$ & Coll. (\%) $\downarrow$ & F1 $\uparrow$ & Lat. $\downarrow$ & Ch. $\downarrow$ \\ 
        \midrule
        \textbf{Scene Generation}: \\
        3$s$ & 0.673 & 1.862 & 1.570 & 0.594 & 1.826 & 1.030 & 0.762 & 0.220 & 5.552 \\ 
        5$s$ & 1.808 & 5.241 & 4.232 & 1.041 & 3.792 & 1.540 & 0.640 & 0.211 & 14.05 \\ 
        8$s$ & 4.646 & 13.86 & 7.906 & 1.834 & 7.130 & 2.325 & 0.522 & 0.214 & 27.61 \\ \midrule
        \textbf{Traffic Simulation}: \\
        3$s$ & 0.720 & 1.918 & 1.257 & 0.701 & 2,013 & 0.857 & - & - & - \\ 
        5$s$ & 1.768 & 4.736 & 2.932 & 1.169 & 4,019 & 1.250 & - & - & - \\ 
        8$s$ & 3.844 & 9.850 & 4.915 & 2.002 & 7.325 & 1.830 & - & - & - \\ 
      \midrule
        \textbf{Close-Loop Simulation}: \\
        3$s$ & - & - & - & 0.624 & 2.350 & 1.070 & - & - & - \\ 
        5$s$ & - & - & - & 1.120 & 4.014 & 1.640 & - & - & - \\ 
        8$s$ & - & - & - & 2.018 & 7.166 & 2.379 & - & - & - \\ 
     \midrule
        \textbf{Motion Planning}: \\
        3$s$ & 1.066 & 2.821 & 0.870 & - & - & - & - & - & - \\ 
        5$s$ & 2.544 & 6.649 & 2.402 & - & - & - & - & - & - \\ 
        8$s$ & 5.448 & 14.18 & 3.814 & - & - & - & - & - & - \\ 
        \bottomrule
    \end{tabular}
    \vspace{-4mm}
\label{tab:testhard_main}
\end{table*}

\section{Additional Implementation Details} \label{app:imple}

Our training process consists of two stages: first, training a Map Tokenizer to encode single-frame images, and then training the overall Generative Pre-training for Driving (GPD-1) model.

\textbf{Map Vector Rasterization.} We model the map centerline within a $64 \times 64$ m rectangular region centered on the ego vehicle. The map lines in this region are rasterized onto a $256 \times 256$ pixel canvas, resulting in a binary (0/1) image. We employ ResNet-50~\cite{he2015deepresiduallearningimage} as the image encoder. For the Vector Quantized Variational Autoencoder (VQVAE)~\cite{vq-vae2}, the codebook size is set to 128, and the latent channel dimension is also 128. This encodes the map image into tokens of shape $H \times W \times C = 8 \times 8 \times 128$. The decoder and ground truth (GT) design follow the encoding strategy of SLEDGE~\cite{chitta2024sledgesynthesizingdrivingenvironments}.

We use the 100M-scene dataset from PlanTF~\cite{cheng2023rethinkingimitationbasedplannerautonomous} for both training and validation. During Map Tokenizer training, a random frame is sampled from each scene. Training is performed on 24 NVIDIA A800 GPUs with 80 GB memory over 41 hours, for 1000 epochs. The batch size is set to 64. The AdamW optimizer is employed with a weight decay of 0. The learning rate for the VQVAE codebook vectors is set to $1.5 \times 10^{-3}$, while the rest of the parameters use a learning rate of $3 \times 10^{-4}$. A cosine annealing schedule is applied, with a warmup period of 50 epochs.

\textbf{Agent Multi-Level Quantization Tokenization.} For the position component, we set the quantization intervals $\{s_1, s_2, \ldots, s_N\}$ to $\{1, 0.01\}$. For the heading component, the quantization intervals are set to $\{20, 1\}$.

For the Transformer decoder, we set the dimension to 128, with 6 layers and 8 attention heads. In the final decoding tokens, for the agent, we directly decode $x$, $y$, heading, and visibility, aligning them with the ground truth (GT). For the map, we decode 128 codebook tokens, determining the corresponding indices for each and treating it as a classification problem.

Training is conducted on 8 NVIDIA A800 GPUs for 148 hours, over 20 epochs. The batch size is set to 2. The AdamW optimizer is used with a weight decay of $1 \times 10^{-4}$. The learning rate is set to $1 \times 10^{-3}$, with a cosine annealing schedule and a warmup period of 3 epochs.

\begin{table}[t]
\caption{\textbf{Performance on the map prediction task in the Test14-hard setting.} 
}\vspace{-3mm}
\centering
\setlength{\tabcolsep}{0.05\linewidth}
\begin{tabular}{l|ccc}
\toprule
Duration  & F1 $\uparrow$   & Lat. $\downarrow$  & Ch. $\downarrow$   \\
\toprule
w/ agents and ego\\
3$s$    & 0.922 & 0.188 & 0.673 \\
5$s$     & 0.875 & 0.211 & 1.480 \\
8$s$     & 0.786 & 0.245 & 4.031 \\
\toprule
w/ ego \\
3$s$     & 0.857 & 0.217 & 1.419 \\
5$s$    & 0.793 & 0.241 & 2.971 \\
8$s$     & 0.717 & 0.276 & 6.258 \\
\bottomrule
\end{tabular}
\label{tab:testhardmap}
\vspace{-7mm}
\end{table}

\begin{figure*}[t]
\centering
\includegraphics[width=1.02\textwidth]{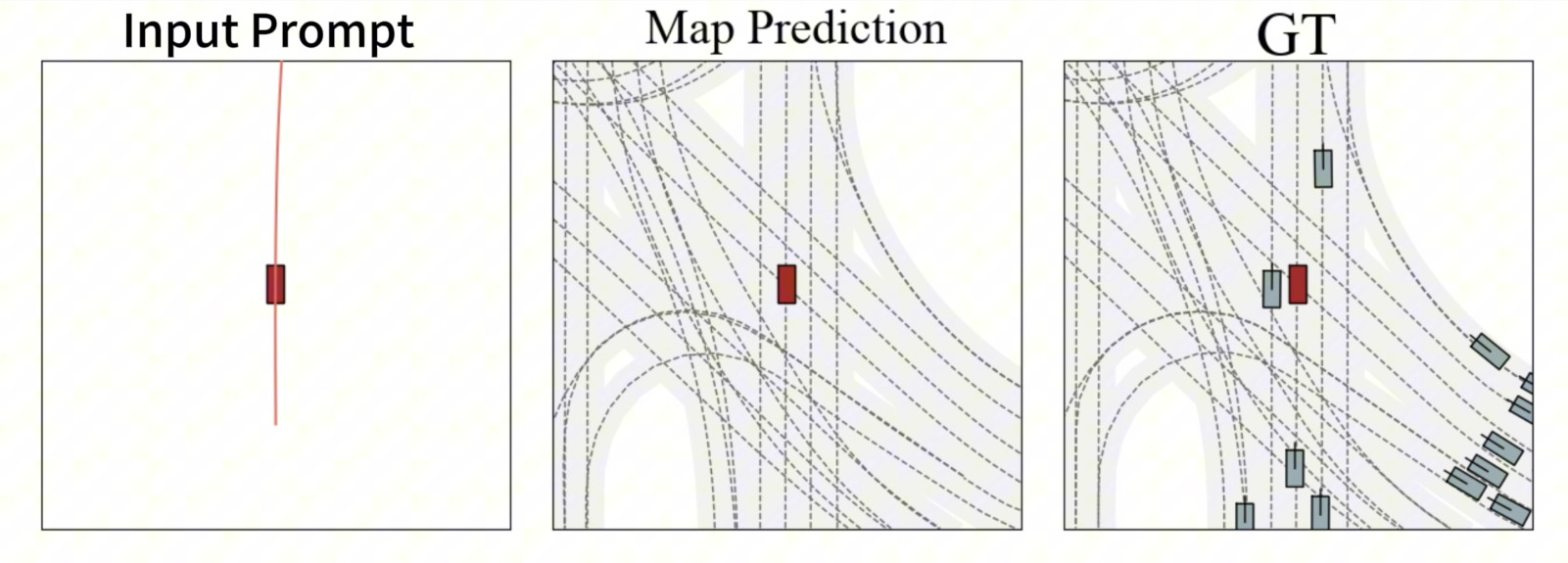}
\vspace{-7mm}
\caption{Sampled images from the video demonstration showcasing the application of the GPD-1 model within the nuPlan framework for Map Prediction tasks.
}
\label{fig:supp}
\vspace{-4mm}
\end{figure*}

\section{Additional Evaluation Metric Details}
\label{sec:addMetrics}
In this paper, we evaluate the performance of agent and ego generation using three metrics: \textbf{Average Displacement Error (ADE)}, \textbf{Final Displacement Error (FDE)}, and \textbf{Collision Rate (Coll.)}. For map generation, we adopt \textbf{F1 Score (F1)}, \textbf{Lateral L2 Distance (Lat.)}, and \textbf{Chamfer Distance (Ch.)}. 

\textbf{Metrics for Agent and Ego Evaluation.}
The metrics used to evaluate the performance of agents and ego are commonly adopted in several previous studies \cite{zheng2024genad, uniad}. Unlike these works, we evaluate not only the trajectory of the ego vehicle but also consider the ego as a special type of agent. The metrics are defined as follows:
\begin{enumerate}
    \item \textbf{ADE}: This measures the L2 distance between the predicted trajectories and the ground truth (GT). In this paper, we focus exclusively on the position of trajectories, ignoring any errors related to heading angles.
    \item \textbf{FDE}: This calculates the L2 distance between the final point of the predicted trajectory and the corresponding point in the GT. For agents, the final point is defined as the last frame in which the agent remains within the visible radius of the ego vehicle.
    \item \textbf{Coll.}: Collision is defined as the intersection of bounding boxes between agents. The collision rate represents the proportion of agents that experienced at least one collision during the evaluation period relative to the total number of interacting agents in the scene.
\end{enumerate}

\textbf{Metrics for Map Evaluation.}
For evaluating map generation, we follow existing methods \cite{chitta2024sledgesynthesizingdrivingenvironments} and employ the following metrics:
\begin{enumerate}
    \item \textbf{F1}: This measures the harmonic mean of precision and recall. Points matched with an error below 1.5m using the Hungarian algorithm are classified as positives, while others are treated as negatives. F1 provides a comprehensive evaluation of the overall similarity between the generated map and the GT.
    \item \textbf{Lat.}: This computes the distance from generated points to their nearest lines in the GT. It offers a detailed assessment of point-level errors in the generated map.
    \item \textbf{Ch.}: The Chamfer Distance measures the mean squared distance from GT points to predicted points and vice versa. This metric evaluates both global consistency and local details of the generated map.
\end{enumerate}

These metrics collectively provide a robust framework for assessing the quality of agent, ego, and map generation in scene simulations.

\section{Additional Results}
\label{sec:addRes}   
In Section 4, we present the results of different settings under the Test14-random \cite{cheng2023rethinkingimitationbasedplannerautonomous} scenarios. For the Test14-hard scenarios, we report the results of Scene Generation, Traffic Simulation, Close-Loop Simulation, and Motion Planning, as shown in Table~\ref{tab:testhard_main}. We observe that the metrics across different settings show a slight decrease, demonstrating the strong generalization ability of our model. 

As shown in Table~\ref{tab:testhardmap}, we present the results of map prediction under the Test14-hard scenarios. It can be observed that using agents and the ego as GT inputs achieves better overall performance compared to using only the ego while other agents remain invisible. This observation is contrary to the conclusion drawn in Test14-random. We believe this is because the hard scenario introduces more complex environments, such as curves and intersections, where the map prediction benefits from the positional information of other agents. However, in simpler scenarios, such as Test14-random, the information from other agents might interfere with map generation.

\section{Video Demonstration}
\label{sec:videoDemo}

Figures~\ref{fig:supp} shows sampled images from the video demo illustrating the application of the GPD-1 model on the nuPlan~\cite{caesar2022nuplanclosedloopmlbasedplanning} validation set. In the accompanying video, we demonstrate the performance of GPD-1 across five different tasks, highlighting the effectiveness of our proposed model.

\end{document}